\documentclass[orivec, runningheads]{llncs}
\usepackage[T1]{fontenc}
\usepackage{graphicx}
\usepackage{cite}
\usepackage{svg}
\usepackage{amssymb}
\usepackage{amsmath} 
\usepackage{adjustbox}
\usepackage{booktabs} % For better looking tables
\usepackage{siunitx}  % For formatting numbers
\usepackage{appendix}
\usepackage{floatrow}
\newfloatcommand{capbtabbox}{table}[][\FBwidth]
\usepackage{caption}
\usepackage{subcaption}
\usepackage{blindtext}
\usepackage{rotating}
\usepackage{multirow}

\begin{document}
\title{Numerical Literals in Link Prediction: A Critical Examination of Models and Datasets}
\titlerunning{Numerical Literals in LP: A Critical Examination of Models and Datasets}

\author{Moritz Blum\inst{1}\orcidID{0000-0003-4924-3903} \and
Basil Ell \inst{1,2}\orcidID{0000-0002-8863-3157} \and
Hannes Ill \and
Philipp Cimiano \inst{1}\orcidID{0000-0002-4771-441X}}

\authorrunning{Blum et al.}

\institute{Bielefeld University, CITEC, Inspiration 1, 33619, Bielefeld, Germany\\
\email{\{mblum, bell, cimiano\}@techfak.uni-bielefeld.de}, \email{hannes.ill@tum.de}
\and
University of Oslo, Problemveien 11, 0313 Oslo, Norway\\
\email{basile@ifi.uio.no}\\
}

\maketitle              % typeset the header of the contribution
\begin{abstract}
Link Prediction~(LP) is an essential task over Knowledge Graphs~(KGs), traditionally focussed on using and predicting the relations between entities. Textual entity descriptions have already been shown to be valuable, but models that incorporate numerical literals have shown minor improvements on existing benchmark datasets. It is unclear whether a model is actually better in using numerical literals, or better capable of utilizing the graph structure. This raises doubts about the effectiveness of these methods and about the suitability of the existing benchmark datasets.

We propose a methodology to evaluate LP models that incorporate numerical literals. We propose i) a new synthetic dataset to better understand how well these models use numerical literals and ii) dataset ablations strategies to investigate potential difficulties with the existing datasets. We identify a prevalent trend: many models underutilize literal information and potentially rely on additional parameters for performance gains. Our investigation highlights the need for more extensive evaluations when releasing new models and datasets. 

\end{abstract}

\section{Introduction}

\begin{figure}[!tb]
  \centering
  \includegraphics[width=0.6\textwidth]{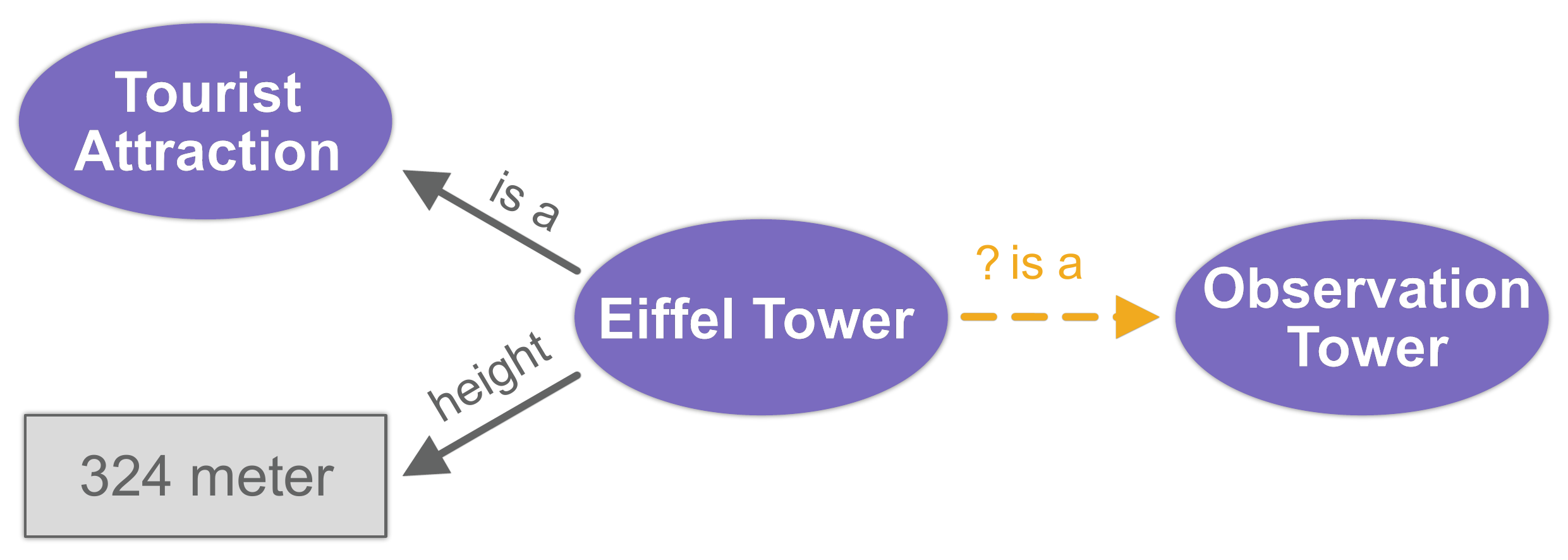}
  \caption{Example KG about the \textit{Eiffel Tower}. The KG contains the entities \textit{Eiffel Tower}, \textit{Tourist Attraction}, and \textit{Observation Tower}; and the literal value \textit{324 meter}: the height of the \textit{Eiffel Tower}. %Ideally, a LP model learned from the data that a \textit{Tourist Attraction} of a certain height (e.\thinspace g., above some threshold) is an \textit{Observation Tower}, to predict: (\textit{Eiffel Tower}, \textit{is a}, \textit{Observation Tower}). 
  }
  \label{fig:kg_example}
\end{figure}

Knowledge Graphs~(KGs) store information in a graph-structured form as sets of relational triples~(i.\thinspace e., triples with a relation that connects two entities) and attributive triples~(i.\thinspace e., triples with a relation that annotates an entity with literal information). Prominent KGs are \textit{Freebase}~\cite{bollacker2008freebase} and \textit{Wikidata}~\cite{vrandevcic2014wikidata}. A small example KG is shown in Fig.~\ref{fig:kg_example}. KGs have emerged as a method to represent and store knowledge in various domains and applications, and will, as the authors believe, play an important role in generative AI as they complement LLMs for Retrieval Augmented Generation~\cite{pan2023llmskgs}. Nevertheless, KGs are inherently incomplete 
%due to the absence of facts 
for various reasons~\cite{10.1145/2623330.2623623}.
%, e.\thinspace g., due to the limitation to certain domains, due to the construction methods, or due to the knowledge source. 
In the past, efforts have been made to develop Link Prediction~(LP) methods to predict missing triples based on the triples already available.

Most LP approaches focus on relational triples, i.\thinspace e., only use relational triples to predict missing relational triples. Neither do these models make use of attributive triples %, i.\thinspace e., relations between entities and values such as integers, strings, etc., 
when predicting relational triples, nor do these models predict attributive triples -- these models ignore information encoded in literals that might be valuable. In the example shown in Fig.~\ref{fig:kg_example}, a model should predict the \textit{is a} relation between \textit{Eiffel Tower} and \textit{Observation Tower}. Ideally, a model has learned from the data that a \textit{Tourist Attraction} that has a certain height (e.g., above some threshold) is an \textit{Observation Tower}. Here, a model needs to incorporate information expressed by the relational triple (\textit{Eiffel Tower}, \textit{is a}, \textit{Tourist Attraction}) and information expressed by the attributive triple (\textit{Eiffel Tower}, \textit{height}, \textit{324 meter}) to predict the missing triple.

To incorporate literals, specialized models~\cite{yao2019kg, wang-etal-2022-simkgc} or extensions of models were proposed~\cite{kristiadi2019incorporating, wu2018transea, GarcaDurn2017KBlrnEL, 10.1145/3132847.3132937, 10.1145/3579051.3579069}. Most models only operate on one, and only some on multiple types of literals. Even though some models are technically able to predict literal values (e.\thinspace g.~\cite{bingcong2022numliteralpred, wu2018transea, pezeshkpour2018embedding}), we focus on the more prominent task of incorporating literals into the prediction of  relational triples. 
%Common literal types are textual literals, numerical literals, values with units of measurement~(e.\thinspace g., 1 kg), or images~\citep{gesese2021survey}. 
Language Models~(LMs) that operate on textual entity descriptions recently set state-of-the-art performance for LP~\cite{saxena2022kgt5}. The inclusion of numerical literals has shown only small improvements in models that do not use literals on common benchmark datasets~\cite{kristiadi2019incorporating, wu2018transea}. However, we assume that numerical literals are highly valuable, especially for scientific KGs about physical experiments~\cite{blum2023} or manufacturing processes~\cite{strotgen2019towards}, which store a large amount of information as numerical data. We focus on numerical literals as there is a lack of research on how well LP models that were designed to be able to use numerical literals can and do make use of numerical literals.

Typically, when new models or model extensions that can incorporate literals are published, then they are compared to state-of-the-art models that can incorporate numerical literals and to the base model they extend
%(as these models are often extensions of traditional LP models that do not incorporate literals) 
by standard metrics like Mean Reciprocal Rank~(MRR) on benchmark datasets. As the improvements through incorporating literals is minor, we can not be sure whether the models are using the attributive triples, or whether the attributive triples in the existing datasets are valuable. 

The benchmark datasets that contain literals are often created by enriching standard LP datasets, e.\thinspace g., FB15k-237 or YAGO3-10, with literal information from their larger source KGs. These datasets might not be perfectly suited for the evaluation of LP models that incorporate literals, as, e.\thinspace g., a certain amount of attributive triples in FB15k-237 connect entities to identifiers (IDs) for other databases that are not exploitable by a model.\footnote{Tab.~\ref{tab:attributive_relations_fb} in App.~\ref{sec:datasetstatistics} shows that for FB15k-237, 3/10 of the triples related to the most frequent attributive relations hold such IDs, overall  6.9\% of all attributive triples in the dataset.} Here, we only point to the ID relations, but other attributive triples might also not be valuable. Something similar might be the case for most of these datasets. To the best of our knowledge, no published evaluation has proven that the numerical literals in these datasets provide information relevant for LP. Therefore, we can not investigate whether the literals are used by the models, making them not suitable for benchmarking.
%The number of datasets explicitly developed for the task of LP with literals is low. One example is LiterallyWikidata~\cite{gesese2021literallywikidata}. As for the enriched traditional LP datasets, it is not ensured that they hold much valuable or new information not already encoded in relational triples. The only difference is that it is ensured during the creation of the dataset that each entity in the dataset is provided with a minimum number of literal values. 

Overall, research on LP with numerical literals lacks a detailed evaluation of and comparison with the existing models and lacks insights on the used benchmark datasets. 
%The current evaluation procedure for LP models that predict relational triples and incorporate attributive triples has drawbacks. 
%If, e.\thinspace g., one model designed with the purpose of being capable of incorporating attributive triples outperforms another model designed with the same purpose, then the superior model is not necessarily better capable of making use of the information represented via attributive triples.
%Instead, the superiorly-performing model could be better capable of making use of the information represented via relational triples due to additional model parameters and even less capable of making use of the information represented via attributive triples.
%Further %
%Complications of model evaluation come from the facts that i) for the specific datasets the models are evaluated on, attributive triples might not be relevant, ii) information can be represented redundantly as attributive and relational triples, and iii) it might be sufficient for a model to make use of the information that an entity has some value for a given relation, where the actual value is irrelevant. 
Therefore, we present the following contributions:
i)~We propose method to extend a dataset with relational and attributive triples, where the prediction of relational triples of that kind can only then be carried out successfully if a model makes use of the attributive triples.
ii)~We propose ablation strategies for the existing evaluation datasets to investigate whether the numerical literals provide any additional knowledge, or whether the numerical literals only add information already contained in the relational triples. % Another setting would be that the literals are just difficult to use, but this can be put into the discussion. 
iii)~We evaluate existing LP models which state to incorporate numerical literals on our semi-synthetic benchmark dataset and on datasets that we obtained by applying our ablation strategies to existing benchmark datasets, to gain insights into the models capabilities to incorporate literals and to gain insights into the suitability and difficulty of the existing datasets. 

\section{Preliminaries}

With $G$ we denote a directed labeled multi-graph with numerical literals. $G$ is a set of triples $(s,p,o) \in \mathcal{U} \times \mathcal{U} \times (\mathcal{U} \cup \mathbb{R})$, where $\mathcal{U}$, and $\mathbb{R}$ are disjoint sets of URIs and numerical values.\footnote{
Although typically KGs contain various types of literals, such as string literals or date literals, here we focus only on numerical literals and we do not distinguish between different types of numerical literals such as integer and float.} The set of triples can be categorized into relational triples $G_E$ and attributive triples $G_A$:
\begin{alignat*}{4}
  &G_E &&= \lbrace (s,p,o) &&~|~ \exists (s,p,o) \in G ~\textrm{ s.t. }~ o \in \mathcal{U} &&\rbrace \\
  &G_A &&= \lbrace (s,p,v) &&~|~ \exists (s,p,v) \in G ~\textrm{ s.t. }~ v \in \mathbb{R} &&\rbrace
\end{alignat*}

    %$$G_E = \lbrace (s,p,o) ~|~ \exists (s,p,o) \in G ~\textrm{ s.t. }~ o \in \mathcal{U} \rbrace$$
    
    %$$G_A = \lbrace (s,p,v) ~|~ \exists (s,p,v) \in G ~\textrm{ s.t. }~ v \in \mathbb{R} \rbrace$$

\noindent The set of entities $\mathcal{E} \subseteq \mathcal{U}$, the set of entity relations $\mathcal{R}_E \subseteq \mathcal{U}$, and the set of attributive relations $\mathcal{R}_A \subseteq \mathcal{U}$ are defined as follows: 
\begin{alignat*}{3}
&\mathcal{E} &&= \lbrace x &&~|~ \exists (s,p,o) \in G ~\textrm{ s.t. }~ x {=} s \lor (x {=} o \land o \in \mathcal{U}) \rbrace\\
&\mathcal{R}_E &&= \lbrace p &&~|~ \exists (s,p,o) \in G_E \rbrace\\
&\mathcal{R}_A &&= \lbrace p &&~|~ \exists (s,p,v) \in G_A \rbrace
\end{alignat*}

    %$$\mathcal{E} = \lbrace x ~|~ \exists (s,p,o) \in G : x {=} s \lor (x {=} o \land o \in \mathcal{U}) \rbrace $$

    %$$\mathcal{R}_E = \lbrace p ~|~ \exists (s,p,o) \in G_E \rbrace \text{, }  \mathcal{R}_A = \lbrace p ~|~ \exists (s,p,v) \in G_A \rbrace$$

%\noindent And it holds that $\mathcal{E} \cap \mathcal{R}_E \cap \mathcal{R}_L = \emptyset$.

%The value in the dimension corresponding to the relation $r \in \mathcal{R}_A$ is sampled from the set $\lbrace v ~|~ (s,r,v) \in G_A\rbrace$. Note that some relations are functional, which means that there exists at most one value per entity and relation. Some relations are not functional. In that case we need to build a set of values and sample one value from that set. 

\subsection{Link Prediction Models}

LP models are trained to predict missing triples using triples already available.

\subsubsection{Traditional Link Prediction Models} 
Traditional LP models can be considered as a function $f$ that assigns a score $f(\Vec{s}, \Vec{p}, \Vec{o}) \in \mathbb{R}$ to each triple $(s, p, o)$ where $s,p,o \in \mathcal{U}$ and $\Vec{e}$ denotes the embedding of the entity $e$. These models are trained to score true triples~(i.\thinspace e., triples in $\mathcal{G}_E$) higher than false triples~(i.\thinspace e., triples not in $\mathcal{G}_E$). % by optimizing a loss function. % $\lambda_E$. 
%Commonly used loss functions are cross-entropy loss and $L_1$ loss. 
Notably, the conventional LP models do not incorporate literals. Popular models are TransE~\cite{bordes2013translating}, DistMult~\cite{yang2015distmult}, ComplEx~\cite{trouillon2016complex}, and TuckER~\cite{balazevic-etal-2019-tucker}. 

%One commonly used model is TransE~\citep{bordes2013translating} that uses the translational scoring function $|| \Vec{s} + \Vec{p} - \Vec{o} ||$, where $\Vec{e} \in \mathbb{R}^d$. In contrast, the commonly used model DistMult~\citep{yang2015distmult} uses a multiplicative scoring function ${<}\Vec{s}, \Vec{p}, \Vec{o}{>} $, where $\Vec{e} \in \mathbb{R}^d$. ComplEx~\citet{trouillon2016complex} extend \textit{DistMult} to embedding vectors with complex values such that $\Vec{e} \in \mathbb{C}^d$. The transfer into the complex embedding space leads to an increased performance in modeling asymmetric relationships. 

\subsubsection{Link Prediction Approaches Incorporating Numerical Literals} 

Some LP models that are able to incorporate numerical literals are extensions of traditional LP models. These models use a feature vector $\Vec{x}_e$ for each entity $e \in \mathcal{E}$. Each dimension of the feature vector $\Vec{x}_e$ corresponds to a relation $r \in \mathcal{R}_A$. The value for a dimension is randomly selected from $\lbrace v ~|~ \exists (e,r,v) \in G_A\rbrace$ or is set to \textit{"0"}, when the entity $e$ has no value for the relation $r$ in $G_A$.\footnote{Replacing non-existing features with \textit{"0"} can be considered critical as \textit{"0"} might also be a valid literal value. Established methods like, e.\thinspace g., LiteralE, make this abstraction. To ensure no negative effect, we computed the proportion of \textit{"0"} literal values in the used datasets which is marginally small: FB15k-237 0.006\%, YAGO3-10 0\%, LitWD48K 1.67\%.} These models can be categorized into two types. 

\paragraph{Fusion via a modification of the scoring function} Such models use the numerical features $\Vec{x}_e$ as (additional) input features, i.\thinspace e., they modify the scoring function explicitly. We investigate two established approaches: 
i)~\textit{LiteralE}~\cite{kristiadi2019incorporating}, by Kristiadi et al., extends traditional LP models by adding a learnable parametric gate function $g(\Vec{e},  \Vec{x}_e)$ to obtain a literal-enriched entity embedding that replaces the initial embedding in the scoring function. This makes LiteralE universally combinable with most existing embedding methods. In this paper, we evaluate LiteralE$_{DistMult}$ and LiteralE$_{ComplEx}$. 
ii)~\textit{KBLN}~\cite{GarcaDurn2017KBlrnEL} is a reduced variant of KBLRN by Garc{\'{\i}}a{-}Dur{\'{a}}n et al. KBLRN is a product of experts model that combines relational~(vectors that describe in which graph patterns an entity occurs), latent~(entity and relation embeddings), and numerical literal features. KBLN leaves out the expert for relational information.

\paragraph{Fusion via a modification of the objective function} Such models learn to predict numerical features jointly with the LP objective.\footnote{The literal information is implicitly encoded into the embeddings and not explicitly provided during inference.} 
Thereby, the entity embeddings incorporate information from both the graph structure and numerical literals. In this paper, we investigate two established approaches: 
i)~\textit{MTKGNN}~\cite{tay2017mtkgnn}, by Tay et al., introduces a neural network for numerical value regression in addition to a neural network for triple scoring. 
ii)~\textit{TransEA}~\cite{wu2018transea}, by Wu et al., is an extension of TransE~\cite{bordes2013translating} that learns a set of functions $g = \lbrace g_p ~|~ p \in \mathcal{R}_A \rbrace$ for numerical value regression.\\

A different line of research investigates methods applied to the datasets instead of the models.

\paragraph{Fusion via literal transformations} Models that transform attributive triples into relational triples allow traditional LP models to incorporate literal information without modifying the scoring function or objective function. In this paper, we investigate the following approaches considered state-of-the-art in LP with numerical literals: \textit{KGA}\cite{ijcai2022p316}, by Wang et al., transforms numerical attributive triples into relational triples by discretizing numerical values into bins, and chaining these bins, modeling multiple levels of granularity. \\

For a broader overview of literal-aware LP models, we refer to \cite{gesese2021survey}.

\subsection{Datasets}
\label{datasets}

Widely used LP datasets that contain numerical literals are: FB15k-237~\cite{toutanova2015fb15k-237}, YAGO3-10~\cite{nickel2012factorizing}, and LitWD48K~\cite{gesese2021literallywikidata}. An extensive overview about existing datasets and their types of literals can be found in \cite{gesese2021literallywikidata}. 
%We have developed a method to create synthetic extensions of non-synthetic datasets and applied the method to FB15k-237. 

We briefly describe the datasets used in this paper. These datasets are publicly accessible under \textit{CC-BY} licenses. % employed in our experiments:
i)~\textit{FB15k-237} is a subset of FB15k which is a subset of Freebase. Toutanova et al. created FB15k-237 by removing inverse relations 
%\footnote{Inverse relations are relationships where if a connection exists between two entities in one direction, there is also a corresponding connection in the opposite direction.} 
from FB15k that allowed even simple models to achieve high scores by simply inverting triples~\cite{toutanova2015fb15k-237}. We use the version of FB15k-237 that was extended with numerical literal as provided by Kristiadi et al.~\cite{kristiadi2019incorporating}.\footnote{See \url{https://github.com/SmartDataAnalytics/LiteralE/tree/master/data}.}
ii)~\textit{YAGO3-10} is a subset of \textit{YAGO3}~\cite{mahdisoltani2014yago3} that only contains triples associated with entities that occur in at least ten relations leading mostly to triples related to people. YAGO3-10 does not contain literals, but they can be derived from \textit{YAGO3}. Again, we use the numerical literals provided by Kristiadi et al.
iii)~\textit{LiterallyWikidata}~\cite{gesese2021literallywikidata} comprises three datasets designed to evaluate LP models utilizing literal data, sourced from \textit{Wikidata} and \textit{Wikipedia}. These datasets vary in size and structure; we use the largest, LitWD48K.

Tab.~\ref{tab:datasetstatistics} in App.~\ref{sec:datasetstatistics} shows the general characteristics of the  datasets used in this paper. 
%These datasets vary in size and structure. %, with triples containing numerical literals of type \texttt{xsd:decimal} and \texttt{xsd:float}. 
%We use the largest dataset: LitWD48K. %Given that \texttt{xsd:float} literals in the dataset predominantly contain geographical coordinates, we focused our experiments on \texttt{xsd:decimal} literals for simplicity. 

\subsection{Evaluation Metrics}
\label{evaluation_metrics}

The models are compared via the filtered mean rank~(MR) metric, the mean reciprocal rank~(MRR) metric, and Hits@k for $k \in \{1,3,10\}$, as proposed by \cite{bordes2013translating}.
For each triple in the test set, the subject and the object entities are corrupted by replacing them by any $e \in \mathcal{E}$. The score for each triple is used to rank the test triple among all of those triples by sorting in ascending order. Triples already contained in the graph are removed before ranking, to not cause true triples to increase the rank of the test triples. 

%The MR is the mean of all computed ranks, the MRR is the mean of the multiplicative inverse of all computed ranks, and the Hits@k are the proportion of ranks that are less or equal to $k$. 

The MR is the mean rank, the MRR is the mean of the multiplicative inverse of the ranks, and Hits@k is the proportion of ranks $\leq k$.

%As a result, the MR is the mean of all computed ranks $MR=\frac{1}{|I|} \sum\nolimits_{r \in I} r$, the MRR is the mean of the multiplicative inverse of all computed ranks $MRR=\frac{1}{|I|} \sum\nolimits_{r \in I} r^{-1}$, and the Hits@k are defined as: $ Hits@k=|\{r \in I ~|~ r \leq k\}| $, where $I$ is the set of all computed ranks. 

\section{Related Work}

To the best of our knowledge, no existing work focuses on a methodology for evaluating LP models with numerical literals or other types of literals.

LP models are evaluated according to standard metrics such as \textit{MR} and \textit{MRR}, following the evaluation protocol proposed by Bordes~\cite{bordes2013translating}, where models are treated as black-boxes. Safavi et al. raise concerns about the reliability of these ranking-based metrics~\cite{safavi2020evaluatinglp}. They point out that while a ranking metric may suggest good performance because the correct triple is ranked high, it could still receive a lower score than an incorrectly top-ranked triple.

In-depth evaluations, e.\thinspace g., analyzing particular relations or distinguishing between head and tail predictions (as done by Bordes et al.~\cite{bordes2013translating}), are uncommon. Such an analysis can be useful to investigate how specific KG characteristics can be learned by a model, e.\thinspace g., if symmetric relations can be properly represented by the model.  %We are not aware of any such method to investigate how literal values are incorporated by a model.

%Moreover, analyzing single predictions is challenging due to variations in model predictions across multiple training runs, making case studies, e.\thinspace g., analyzing individual predictions, inaccurate. 

Explainability methods could provide insights into the behavior of LP models. 
Whereas rule-based approaches~\cite{nassiri2023regnum} and explainers over graph neural network-based approaches~\cite{zhang2023page, ying2019gnnexplainer} can offer explanations for model predictions, particularly shallow models like TransE or DistMult are difficult to explain. Ismaeil et al. %introduce a method to 
generate interpretable vectors for entity embeddings~\cite{youmna2023feabi}. They employ embedded feature selection techniques to extract propositional features from the KG that are important for a given KG embedding model.

Another way to gain insights into the specific behavior and capabilities of LP models is to build datasets in a way such that obtaining good LP results requires the models to have specific capabilities, thus these datasets enable to test to what extent a model has these capabilities. E.\thinspace g., recent work stated a lack of evaluation datasets covering certain KG properties like a given entity type system, pairs of mutually inverse relations, or mediator objects to represent n-ary relationships. Shirvani-Mahdavi et al. evaluate these properties on a newly proposed version of the Freebase KG~\cite{shirvani2023freebaselp}.

The outlined open challenge of evaluating and explaining traditional LP models without considering literals results in a scarcity of research on the evaluation and explainability of LP with literals. 

Although the original releases of existing LP datasets such as FB15k-237 and YAGO3-10 lack attributive triples, they have been extended with textual and numerical attributes sourced from their respective KGs. %, Freebase and YAGO3~\cite{kristiadi2019incorporating}. 
However, these enriched datasets contain numerous entities without numerical attributes,\footnote{See Tab.~\ref{tab:datasetstatistics} in App.~\ref{sec:datasetstatistics}} and the provided numerical attributes are not proven to be helpful for LP, as, e.\thinspace g.,  6.9\% of the attributive triples in FB15k-237 hold IDs. Consequently, Gesese et al. introduced a series of LP datasets called LiterallyWikidata, constructed from Wikidata and Wikipedia, specifically for LP involving numerical and textual literals~\cite{gesese2021literallywikidata}. The graph structure of LiterallyWikidata was designed for benchmarking LP models, avoiding issues such as that inverse relations could leak information or the existence of any shortcut features.
%skewed relations mostly pointing to a single head or tail entity. 

Despite the existence of datasets tailored for LP tasks involving (numerical) literals, these datasets are derived from real KGs, making it challenging to accurately assess the true advantages of integrating (numerical) literal information. 

%The most related to our work is the input feature ablation study done by \citet{GarcaDurn2017KBlrnEL} to investigate which graph structure features positively contributed to the performance of their model.

The most related work is Garc{\'{\i}}a{-}Dur{\'{a}}n et al.'s input feature ablation study, which investigated which graph structure features improved their model's performance~\cite{GarcaDurn2017KBlrnEL}.

%As the existing evaluations are limited in their expressively, there is a need for such evaluation methodologies. Moreover, while synthetic benchmarking datasets like those proposed for graph classification tasks by Dwivedi et al.~\cite{dwivedi2023benchmarking} exist, there is a notable absence of such synthetic datasets for investigating the role of literals in LP.

\section{Methodology}

We i) propose a method to enrich an existing dataset with synthetic information that enables us to find out if numerical literal-aware models are capable of using numerical literals to make predictions about relational triples. Furthermore, ii) we develop a set of ablation methods to gain further insights into the existing literal-aware datasets, whether in some datasets attributive triples might not be used for LP, or whether information is represented redundantly as attributive and relational triples.

%Given that most existing models only show a small benefit from extending a base models to numerical literal-aware models, doubts arise regarding the effectiveness of these models on established benchmark datasets.

%We argue that a deeper investigation could reveal interesting properties of the models. Therefore, 

%We define the following properties to describe potential reasons for model performances:

%\begin{description}
%\item[Attributive Capability] (short: capA) -- the model is capable of making use of information represented via attributive triples

%\item[Redundancy] (short: redAR) -- information is redundantly represented in the form of attributive triples and relational triples

%\item[Attributive Relevance] (short: relA) -- information represented in the form of attributive triples is relevant for link prediction

%\item[Existing Relevance] (short: existSuff) -- information about the existence of a value is sufficient
%\end{description}

When elucidating the derivable conclusions from the following ablation experiments, we denote a model trained on the dataset $D$ as $m(D)$ and define $\sigma(m(D))$ as the result of evaluating $m(D)$ according to some measure of performance $\sigma$ (such that a higher value indicates better performance).

\subsection{Semi-Synthetic LP Dataset with Literals}
\label{synthetic_dataset}

To ensure attributive triples to be relevant, we propose a dataset extension methodology
%. Given a dataset, we remove all attributive triples and add new ones. Then we add new relational triples that connect some of the entities already contained in the graph to new entities, 
with the intention to introduce a new learning goal given by a function $h$ into the dataset.

For simplicity, we restrict $h$ to be a function that predicts a relation $r_{syn{-}r}$ from an existing entity $e$ to one of two classes added to the dataset, namely $c_{high}$ and $c_{low}$, based on the attributive triple $(e, r_{syn-a}, v)$. More precisely, our function $h$ is defined as:
%$$
%h(e) = \begin{cases} 
%        (e, r_{syn{-}r}, c_{high}) & \text{if } (e, r_{syn{-}a}, v) \in G' \text{ with }  v > 0.5 \\
%        (e, r_{syn{-}r}, c_{low}) & \text{if }  (e, r_{syn{-}a}, v) \in G' \text{ with }  v \leq 0.5
%        \end{cases}
%$$
$$
h(e) = \begin{cases} 
        \begin{aligned}
        (e, r_{\text{syn-}r}, c_{\text{high}}) & \quad \text{if } \exists (e, r_{\text{syn-}a}, v) \in G' 
        & \quad \text{with } v > 0.5 \\
        (e, r_{\text{syn-}r}, c_{\text{low}}) & \quad \text{if }  \exists  (e, r_{\text{syn-}a}, v) \in G' 
        & \quad \text{with } v \leq 0.5
        \end{aligned}
        \end{cases}
$$

\begin{figure}[!tb]
  \centering
  \includegraphics[width=0.6\textwidth]{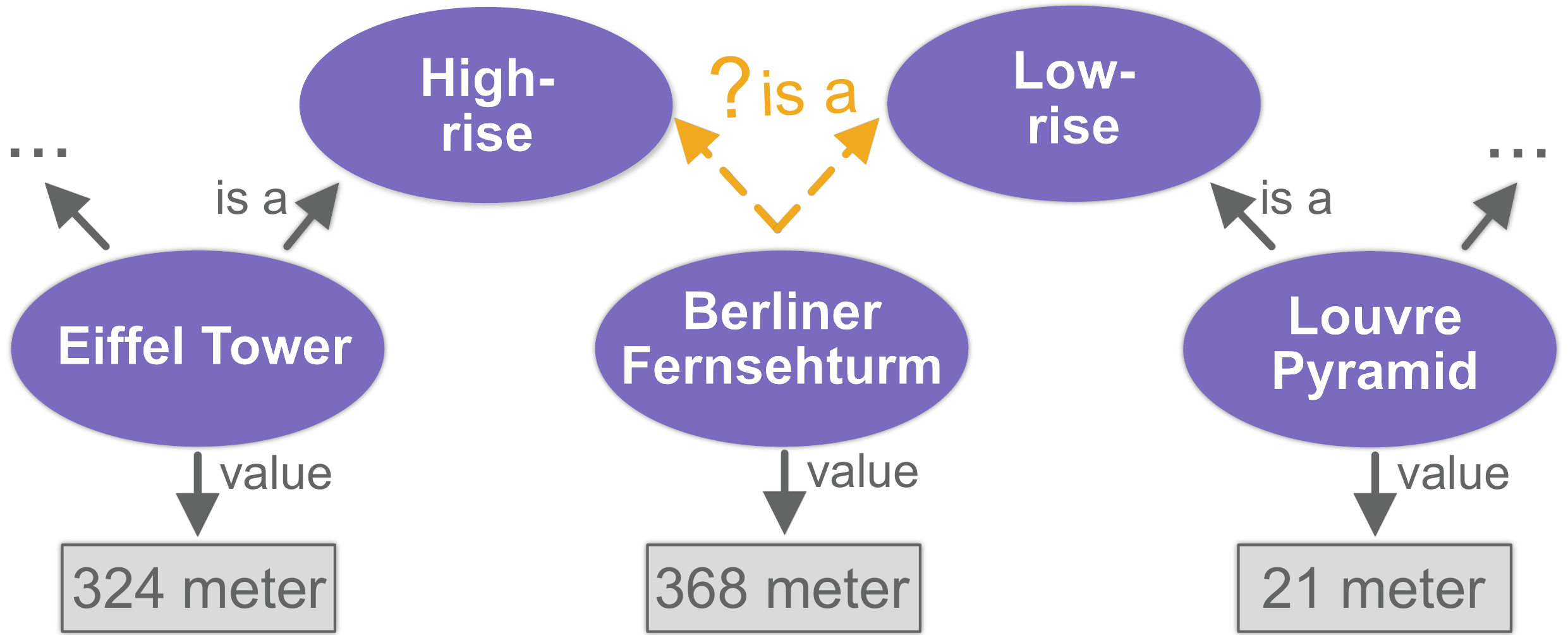}
  \caption{Example of the synthetic dataset enrichment. The entitites \textit{High-rise} and \textit{Low-rise} represent $c_{high}$ and $c_{low}$ and \textit{is a} is used as the $r_{syn-a}$ relation. Ideally, an LP model predicts the tail entity \textit{High-rise} for the given head \textit{Berliner Fernsehturm} and the \textit{is a} relation.
  }
  \label{fig:synthetic_example}
\end{figure}

To remove any noise from the original $G_A$, we replace it by $G_A'$ defined as $\lbrace (e, r_{syn{-}a}, v) ~|~ e \in \mathcal{E'} \rbrace$ where $v {\sim} \text{Uniform}(0,1)$ and $\mathcal{E'} \subseteq \mathcal{E}$. We then apply $h$ to every $e \in \mathcal{E'}$ to obtain relational triples that are added to $G_E'$. The new dataset is defined as $G' = G_E' \cup \mathcal G_A'$. An example is shown in Fig.~\ref{fig:synthetic_example}.

Note that the function $h$ could be more complex, taking into account multiple relational and attributive triples, make use of more than the one target relation $r_{syn{-}r}$ and the two target entities $c_{high}$ and $c_{low}$, and realize something more complex than comparing a value against a threshold value.

%More complex $h(e)$ would take into account $G_A'$ as well as $G_E$. However, for simplicity, we restricted to the most simple case. 

%XXXXXXXX this generates training data. do we also need to describe that test data only contains the syn-a attributive triples, because the task should consist in predicting the syn-r triples?

Let $\mathcal{E}_{high}$ and $\mathcal{E}_{low}$ be defined as follows:
$\mathcal{E}_{high} := \lbrace e \in \mathcal{E}' ~|~ \exists (e,r_{syn{-}a}, v) \in G_A' \land v > 0.5\rbrace$
and
$\mathcal{E}_{low} := \lbrace e \in \mathcal{E}' ~|~ \exists (e,r_{syn{-}a}, v) \in G_A' \land v \leq 0.5\rbrace$. (Note that $\mathcal{E} = \mathcal{E}_{high} \cup \mathcal{E}_{low}$ and $\mathcal{E}_{high} \cap \mathcal{E}_{low} = \varnothing$.)

Our goal is to measure the models' ability to score the synthetic relational triples according to $h$. As this is a binary classification task, we define the accuracy, denoted by $Acc$, as follows:
%$$Acc = \frac{true_{high} + true_{low}}{|\mathcal{E}_{high}|   +  |\mathcal{E}_{low}|}$$
%\begin{align*}
%Acc &:= \frac{true_{high} + true_{low}}{|\mathcal{E}_{high}|   +  |\mathcal{E}_{low}|}\\
%true_{high} &:= \sum\limits_{e \in \mathcal{E}_{high}}   | \lbrace  r(e,r_{syn{-}r}, c_{high}) \geq r(e, r_{syn{-}r}, c_{low})  \rbrace |\\
%true_{low} &:= \sum\limits_{e \in \mathcal{E}_{low}}   | \lbrace  r(e,r_{syn{-}r}, c_{low}) \geq r(e, r_{syn{-}r}, c_{high})  \rbrace |
%\end{align*}
%\begin{align*}
%Acc &:= \frac{true_{high} + true_{low}}{|\mathcal{E}_{high}| + |\mathcal{E}_{low}|} \\
%true_{high} &:= \sum\limits_{e \in \mathcal{E}_{high}} \left| \left\{ r(e, r_{\text{syn-}r}, c_{\text{high}}) \geq r(e, r_{\text{syn-}r}, c_{\text{low}}) \right\} \right| \\
%true_{low} &:= \sum\limits_{e \in \mathcal{E}_{low}} \left| \left\{ r(e, r_{\text{syn-}r}, c_{\text{low}}) \geq r(e, r_{\text{syn-}r}, c_{\text{high}}) \right\} \right|
%\end{align*}
\begin{equation}
Acc := \frac{true_{high} + true_{low}}{|\mathcal{E}_{high}| + |\mathcal{E}_{low}|}
\end{equation}
where $true_{high}$ is the number of $e \in \mathcal{E}_{heigh}$ for which $r(e, r_{syn-a}, c_{high}) \geq  r(e, r_{syn-a}, c_{low})$. $true_{low}$ is defined analogously.

%$$true_{high} := \sum\limits_{e \in \mathcal{E}_{high}}   | \lbrace  r(e,r_{syn{-}r}, c_{high}) \geq r(e, r_{syn{-}r}, c_{low})  \rbrace |$$

%$$true_{low} := \sum\limits_{e \in \mathcal{E}_{low}}   | \lbrace  r(e,r_{syn{-}r}, c_{low}) \geq r(e, r_{syn{-}r}, c_{high})  \rbrace |$$.

%$$
%Acc = \frac{\sum\limits_{e \in \mathcal{E}_{high}}   | \lbrace  r(e,r_{syn{-}r}, e_{high}) \geq r(e, r_{syn{-}r}, e_{low})  \rbrace | + \sum\limits_{e \in \mathcal{E}_{low}}   | \lbrace  r(e,r_{syn{-}r}, e_{low}) \geq r(e, r_{syn{-}r}, e_{high})  \rbrace |}{   |\mathcal{E}_{high}|   +  |\mathcal{E}_{low}| }
%$$

We consider the following situations where we can derive conclusions:

\noindent i) if $\sigma(m(G_E \cup G_A')) < \sigma(m(G_E \cup G_A))$, i.\thinspace e., the model that has no access to the original attributive triples performs worse, then the attributive triples are used for the prediction;

\noindent ii) if $\sigma(m(G_E \cup G_A')) \geq \sigma(m(G_E \cup G_A))$, i.\thinspace e., both models perform equally well, or the model that used the random features performs better, then the model is not capable of making use of literals. \\

\subsection{Literal Features Ablation}
\label{literal_features_ablation}

If a model has proven to incorporate literals from a synthetic dataset into the prediction, this model should be evaluated on the established benchmark datasets. To gain insights into the performance increase by using literals, one has to compare models that use literals against the same model without access to literals.

For some models, e.g., MTKGNN, we cannot remove the attributive triples from the dataset, as these models directly operate on the numerical features only, and do not learn a separate entity embedding. For other models, e.\thinspace g., LiteralE, removing the attributive triples from the dataset reduces the number of model parameters.

Therefore, we propose an ablation method where each entity is related with each attributive relation to a certain value. Given $G = G_E \cup G_A$, we create $G' = G_E \cup G_A'$ where $G_A'$ is created as follows: for each entity $e \in \mathcal{E}$ and each relation $p \in \mathcal{R}_{A}$, we add the triple $(e,p,v)$ to $G_A'$ where $v$ is a certain literal value we assign.

%We consider the following situations where we can derive conclusions:

%$$p(m(G_E)) \approx p(m(G_E \cup G_A)) \Rightarrow \neg capA \lor \neg relA \lor redAR$$

%If both models perform equally well, then either the model is not capable of making use of literals ($\neg capA$), or literals are not relevant for the prediction task ($\neg relA$), or information represented via attributive triples is redundantly represented via relational triples ($redAR$).

%$$p(m(G_E)) \ll p(m(G_E \cup G_A)) \Rightarrow relA $$

%If the model that has no access to the original attributive triples performs worse, then the literals are relevant for the prediction task ($relA$).\\

%For some models, e.g., XXXXX, we cannot remove the attributive triples from the dataset and we cannot insert zero values, because these models are only trained with attributive triples.

As some models only operate on numerical features, assigning  the same value to all attributive triples would lead to identical features for all entities. Consequently, such models would lose the ability to distinguish the entities. Therefore, we propose to sample $v$ randomly from $\text{Uniform}(0,1)$.  
%This ablation method can then be used for both kinds of models: those that need both relational and attributive triples, and those that only work with attributive triples.

We consider the following situations where we can derive conclusions about a model and a dataset, under the assumption that the model can make use of literals according to the experiments of the model on the semi-synthetic dataset:

%$$p(m(G_E \cup G_A')) \ll p(m(G_E \cup G_A)) \Rightarrow relA $$
%$$p(m(G_E \cup G_A')) \ll p(m(G_E \cup G_A))$$

\noindent i) if $\sigma(m(G_E \cup G_A')) < \sigma(m(G_E \cup G_A))$, i.\thinspace e., the model that has no access to the original attributive triples performs worse, then the attributive triples are relevant for the prediction task;

%$$p(m(G_E \cup G_A')) \approx p(m(G_E \cup G_A)) \Rightarrow \neg capA \lor \neg relA \lor redAR$$

\noindent ii) if $\sigma(m(G_E \cup G_A')) \geq \sigma(m(G_E \cup G_A))$, i.\thinspace e., both models perform equally well, or the model that used the random features performs better, then, either the information represented via attributive triples is redundantly represented via relational triples, the literal information is difficult to use by the models, or no information relevant for LP is represented via attributive triples.

\subsection{Relational Features Ablation}
\label{relational_features_ablation}

The previous experiments may leave open whether attributive triples are not relevant for the prediction task, challenging to leverage, or whether information is represented redundantly as relational and attributive triples. To gain insights into the redundancy of relational and attributive triples for a given dataset, we propose an ablation method that targets relational attributes, thus modifies $G_E$. We reduce $G_E$ to $G_{E_{-\alpha}}$ s.\thinspace t.

1) $|G_{E_{-\alpha}}| = (1-\alpha) |G_E|$ where $\alpha \in [0,1]$ is a user-defined real value.

2) $\forall e \in \mathcal{E}: (\exists p,o: (e,p,o) \in G_{E_{-\alpha}}) \lor (\exists s,p: (s,p,o) \in G_{E_{-\alpha}})$

3) $\forall p \in \mathcal{R_E}: (\exists s,o: (s,p,o) \in G_{E_{-\alpha}})$

This means, we remove relational triples from $G$ until $|G_{E_{-\alpha}}| = (1-\alpha) |G_E|$. We ensure that there remains at least one triple per entity $e \in \mathcal{E}$ and relation $p \in \mathcal{R_E}$ such that embeddings are learned. Note that for some $G_E$ and $\alpha \in \mathbb{R}^{+}$  it can be the case that there is no $G_{E_{-\alpha}}$ that satisfies both constraints. Thus, there is a limit to how much $G_E$ can be reduced.

We consider the following situations where we can derive conclusions:

\noindent i) if with the reduced set of relational triples the random feature ablation has an effect on model performance (i.e., $\sigma(m(G_{E_{-\alpha}} \cup G_A)) \gg \sigma(m(G_{E_{-\alpha}} \cup G_A'))$), then that means that information is represented redundantly as attributive and relational triples and that attributive triples are relevant for the prediction task;

%$$p(m(G_E \cup G_A')) \approx p(m(G_E \cup G_A)) \land p(m(G_{E_{-\alpha}} \cup G_A)) \gg p(m(G_{E_{-\alpha}} \cup G_A'))$$

\noindent ii) if with the reduced set of relational triples the random feature ablation has still no effect on the model performance (i.e., $\sigma(m(G_{E_{-\alpha}} \cup G_A)) \approx \sigma(m(G_{E_{-\alpha}} \cup G_A'))$), then attributive triples are either difficult to incorporate or not relevant for the prediction task.

%TODO-XXXXXXXXXXX If previous attributive literal ablation studies that literals have no effect on the model performances and $p(m(G_{E_{-\alpha}} \cup G_A)) \geq p(m(G_E \cup G_A))$, then the reduction cannot be compensated via other (attributive) triples, or the model is not capable of exploiting attributive triples and thereby not capable of exploiting the redundancy.

\section{Experimental Setup}
We apply our methodology to all numerical literal-aware models mentioned in~\cite{gesese2021survey} and the state-of-the-art model KG~\cite{ijcai2022p316}. 

\paragraph{Implementation} We run our experiments with LiteralE$_{DistMult}$, LiteralE$_{ComplEx}$, KBLN, and MTKGNN with the code of Kristiadi et al.~\cite{kristiadi2019incorporating}.\footnote{See \url{https://github.com/SmartDataAnalytics/LiteralE}.} 

%We implemented TransEA using PyTorch Geometric,\footnote{See \url{https://pytorch-geometric.readthedocs.io}. We extended the existing TransE implementation to TransEA.} as no implementation was publicly available. 

We implemented TransEA in PyTorch Geometric\footnote{See \url{https://pytorch-geometric.readthedocs.io}. We extended the existing TransE implementation to TransEA.} due to the absence of a public implementation.

%As KGA implements dataset transformations, the approach can use any traditional LP model. 

We decided to use the model variants that achieve the overall best performance and the model that shows the largest performance gains through incorporating literals, which are KGA$_{TuckER}$ and KGA$_{DistMult}$ according to \cite{ijcai2022p316}.\footnote{The KGA transformations approximately create as many additional relational triples as attributive triples. As our proposed attributive features ablation creates a large number of literals with random values, the number of relational triples increases significantly. Consequently, we had to limit the number of attributive relations. Instead of relating each entity with each attributive relation, we only replace the numerical values of attributive triples in the original dataset by a random value. Thereby, the model is provided with some literal information, i.\thinspace e., the existence of the attributive relation.}

All hyperparameters are reported in App.~\ref{sec:hyperparameters}. We ran all experiments three times and computed mean and standard deviation for each metric.

\paragraph{Semi-Synthetic FB15k-237} We apply our dataset enrichment method to the FB15k-237 dataset. We decided that $\mathcal{E}$ is the set of entities of type person. FB15k-237 contains $4,505$ entities of type person, i.\thinspace e. ${\approx}30\%$ of the entities.\footnote{By assigning numerical literals only to certain entities, we enable further analysis, such as determining if the model learns that only certain entities have a specific literal.}  

For evaluation, we create a training, validation, and test split as follows: we add 70\% of the new synthetic relational triples to the original train set, 15\% to the original validation set, and 15\% to the original test set. The new synthetic literal values replace the original literal values. The models are trained for LP as usual.  

\paragraph{Computing Resources} Our evaluation required numerous experiments, due to the combinations of investigated models and datasets. We used 10 A100 GPUs for two weeks. The evaluated models have similar sizes, e.\thinspace g., Literale$_{DistMult}$ trained on FB15k-237 has $\approx$3M parameters. %The models mostly scale linearly with $|\mathcal{E}|$ and quadratic with the number of embedding dimensions. 

%has $|\mathcal{E}| \times d + |\mathcal{R}_E| \times d + 2 \times |\mathcal{E}|^2 + 2 \times |\mathcal{R}_A| \times |\mathcal{E}| + |\mathcal{E}|$  where $d$ is the dimensionality of the embedding vectors.  

\section{Results}

\subsection{Synthetic LP dataset with literals}

We created a semi-synthetic FB15-237 dataset and used it to investigate the models' ability to utilize the necessary numerical literal information for predicting relational triples. The results are shown in Tab.~\ref{tab:synthetic}. The accuracy for all models is shown in the column $Acc_{org}$. A score slightly above 0.5 suggests that the models' performance is only marginally better than random guess, and a score close to 1 suggests that the model is capable of making correct predictions by using the numerical literals.

As a baseline, we train the models with random features following the introduced literal features ablation method, which we applied after the creation of the semi-synthetic dataset. $Acc_{rand}$ is the score of the models when the literals provide no information, forcing the models to guess randomly.

The variance across runs is small; hence, these values indicate a measure of reliable performance.

The KGA models are capable of using the provided numerical literals for their prediction as they achieve $Acc_{org}$'s of 0.999, both. The Acc$_{rand}$ scores range from 0.482 to 0.510, proving the models' random guessing. Note that due to the randomness of the features, the models cannot make a justified prediction. The Acc$_{org}$ scores achieved by the other models are much lower and in the same range as the Acc$_{rand}$ scores', showing that these models do not or do only to a small extent use the information provided via literals.

\begin{table}[]

\setlength{\tabcolsep}{6pt}
\renewcommand{\arraystretch}{1.1}

\caption{Scores achieved on the synthetic dataset. Acc$_{org}$ denotes the Acc score achieved on the synthetic dataset when we provide the meaningful synthetic literal values, whereas Acc$_{rand}$ denotes the Acc score on the synthetic dataset if we apply the random feature ablation after the dataset creation.}
\label{tab:synthetic}
\begin{center}

\begin{tabular}{l|l|l}
\hline
\textbf{Model} &\textbf{ Acc$_{org}$}                            &  \textbf{Acc$_{rand}$}                            \\ \hline
LiteralE$_{DistMult}$ & $0.512 {\scriptstyle \pm 0.003}$ & $0.482 {\scriptstyle \pm 0.001}$\\ 
LiteralE$_{ComplEx}$ & $0.493 {\scriptstyle \pm 0.005}$ & $0.498 {\scriptstyle \pm 0.020}$\\ 
KBLN & $0.482 {\scriptstyle \pm 0.009}$ & $0.493 {\scriptstyle \pm 0.005}$\\ 
MTKGNN & $0.472 {\scriptstyle \pm 0.006}$ & $0.495 {\scriptstyle \pm 0.009}$\\ 
TransEA & $0.489 {\scriptstyle \pm 0.022}$ & $0.496 {\scriptstyle \pm 0.020}$\\ 
KGA$_{TuckER}$ & $0.999 {\scriptstyle \pm 0.000}$ & $0.510 {\scriptstyle \pm 0.007}$\\ 
KGA$_{DistMult}$ & $0.999 {\scriptstyle \pm 0.000}$ & $0.487 {\scriptstyle \pm 0.011}$\\ 

\hline
\end{tabular}

\end{center}
\end{table}

\subsection{Literal Features Ablation}

\begin{figure}[!tb]
  \centering
  \includegraphics[trim={0 0.3cm 0 0}, width=0.8\textwidth]{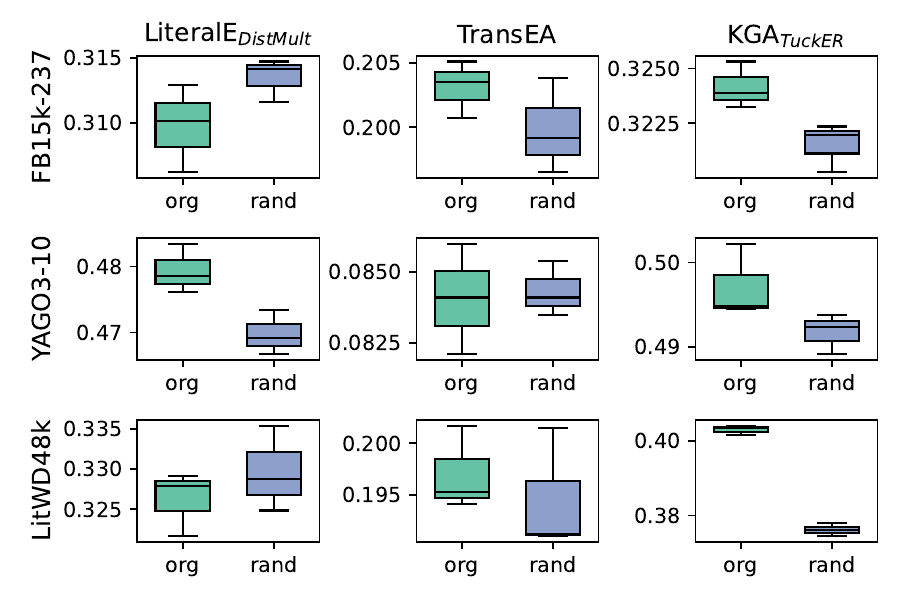}
  \caption{MRR scores over three runs for models and datasets that either include the original literal features or that include random literal features. %Tab.~\ref{tab:org_vs_rand} in App.~\ref{app_tables} shows the MR, MRR, and Hits@10 (mean and variance).
  }
  \label{fig:orgvsrands}
\end{figure}

The results for our experiments for the random literal features ablation method are visualized for three models in Fig.~\ref{fig:orgvsrands} as box plots, showing the mean and variance of the MRR scores across three runs. For each  combination of model and dataset, the plot shows two boxes. The first box shows the score achieved by the model trained with the original literal features and the second box shows the score achieved by the model trained on the random features.\footnote{The box-plots for all experiments and the scores of all computed metrics for these experiments are in App.~\ref{sec:app_tables}.} 

In general, as shown in Fig~\ref{fig:orgvsrands}, replacing original features by random features has no significant negative impact on most of the models as the box for the original features and the box for the random features overlap in many cases and the differences are very small. The detailed reported in App.~\ref{sec:app_tables}, shows that in 9 of the 21 cases we see the models with random literals outperform the models that use real literals regarding the MRR score, and only in 10 of the 15 cases models showed a benefit regarding the MRR score in using the real attributive triples provided by the datasets. The Hits@k scores follow this trend. When looking at the MR scores, this trend exists, too. The KGA models show the best usage of literals as the predictions with the original features are better, but only marginally, than with the random features for all combinations of KGA models and datasets investigated, except for KGA$_{DistMult}$ on the YAGO3-10 dataset.

%As the investigated models do not show a significant benefit in using the real literal features, the conclusions from the other introduced literal ablation strategies for these models on these datasets are limited. Therefore, we only exemplarily show the results for the KBLN and the LiteralE$_{ComplEx}$  model on the FB15k-237 dataset in Tab.~\ref{tab:variations}. The KBLN model shows a drop in MRR from 0.295 to 0.285 when replacing the real literal features by random features. When we replace the literal features and only indicate whether literals exist, % or when we filter the literals by frequency beforehand the MRR only drops to 0.293. The MR and Hits@k scores confirm this trend.

%In contrast, the LiteralE$_{ComplEx}$ shows a small improvement from using the random literal features instead of the real literal features in MRR from 0.272 to 0.278.

%For completeness, we also included the scores for the original ComplEx model without the LiteralE extension into the table. This base model is outperformed by the LiteralE model in terms of MR, however, the MRR and Hits@k scores decrease, thereby, deviating from the results reported by \citet{kristiadi2019incorporating} which reported an improvement according to all metrics.

%\input{tables/variations_tab}

\subsection{Relational Features Ablation}
\begin{figure}[!tb]
  \centering
  \includegraphics[trim={0 0.4cm 0 0}, width=0.8\textwidth]{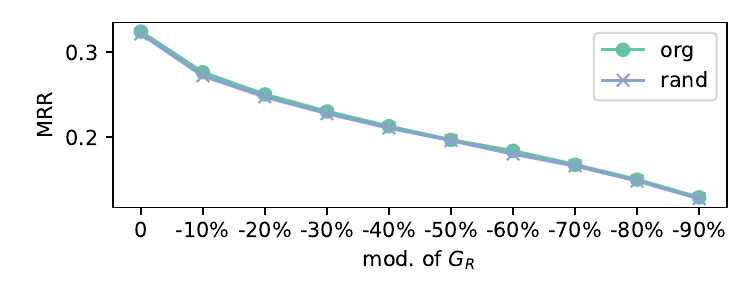}
  \caption{KGA$_{TuckER}$'s MRR scores (mean and variance over three runs) after removing x\% relational triples from FB15k-237. The model is provided either with the original or with random numerical features. The variance is marginally small and not visually recognizable in the figure.}
  \label{fig:rel_abl}
\end{figure}

The ablation effect of relational triples from FB15k-237 on KGA$_{TuckER} $ is shown in Fig.~\ref{fig:rel_abl}.\footnote{Plots for further models are contained in App.~\ref{sec:app_tables}.} We plot the mean and standard deviation of the MRR score while reducing the amount of relational triples from 100\%~(representing the original dataset) to 10\%~(equivalent to removing 90\% of the relational triples) in steps of 10\%. As expected, the MRR score decreases significantly when reducing the available relational triples, showing that important information is eliminated.

We compared these models against the ones we applied the random literal feature ablation strategy on. The curves of the models on random features are very close to the curves of the models on the original features, not showing any advantages of the original features when reducing the relational triples. 

\section{Discussion}

The synthetic dataset creates a scenario where numerical literals are necessary for predictions. KGA converts these continuous literals into discrete entities, allowing models to bypass the need to incorporate the concrete literal values. Thereby, KGA translates the task created by our semi-synthetic dataset into a more simple graph-structure learning task,\footnote{Entities with similar literals obtain similar embeddings as entities with similar literals are connected to the same bins.} but might struggle with more complex synthetic datasets. The other models behaved similar to random guessing, as the scores are close to 0.5. We assume that these model's objective function does not enforce the models to integrate numerical information valuable for LP into the entity embeddings. Even though TransEA forces the embeddings to contain information to reconstruct the numerical literals, this information is not necessarily valuable for LP.

Even though we proposed our synthetic dataset to overcome a certain issue with the existing real-world benchmark datasets, we, nevertheless, believe that real-world datasets are relevant. Therefore, we also evaluated all models on these datasets and compared the scores to the scores of their random feature variant to confirm the previous results. 

% We evaluated LP models that were proposed to incorporate numerical literals on the semi-synthetic dataset where we can be sure that attributive triples are necessary for predicting relational triples. As the resulting scores are close to random guess, we infer that the models do not incorporate the literal information into the prediction of relational triples. %Despite creating a scenario where numerical data is crucial for specific predictions with the synthetic dataset extension, these features are largely underutilized. 
%The models appear to prioritize information from relational triples over available numeric data, struggling to discern the relevance of each type of information in different situations. 

One would expect significant performance drops after applying the random feature ablation methods. We do not observe any significant performance drop, and in some cases even an improvement. Interestingly, the KGA models do only show small benefits of using numerical literals even though they showed good performances on our synthetic dataset. This brings us to the conclusion that either the models are not capable of making use of literals, or the literals are not relevant for the prediction task, or the information contained in attributive triples is difficult to use, or the information represented via the attributive triples is redundantly represented via relational triples.

We are not sure about the reason for the increase in performance after introducing random literals. Possibly, in some cases unintentionally good features are created which can be used by the models. 
A similar increase in performance through random node initialization has been observed by Abboud et al. for GNNs, which gain additional expressivity in the neighborhood encoding from random node initialization~\cite{abboud2021randomfeatures}. However, all models we investigated are shallow models that do not perform any neighborhood encoding. 

We have to note that our scores of the LiteralE models slightly differ from the ones reported by Kristiadi et al.~\cite{kristiadi2019incorporating}, even though we used their implementation and hyperparameters. Interestingly, the base ComplEx model achieves higher MRR and Hits@k scores than the LiteralE$_{ComplEx}$ model on FB15k-237 in our experiments. However, the similarity of the results from three runs confirm our results' reliability.

Lastly, we investigated if relational and attributive triples redundantly represent information, which leads to the LP performance to remain similar even though numerical literals are incorporated. If the attributive triples were redundant, one would expect the model with the real literal features to obtain less worse results when ablating the relational triples than the one with the random features, i.\thinspace e., at some point the literal features should become important. As Fig.~\ref{fig:rel_abl} does not show any benefit of incorporating numerical literals when reducing the amount of available relational triples, we conclude that either the attributive triples are not relevant for the prediction task, or the information contained in attributive triples is difficult to use by the models.

\section{Conclusion \& Future Work}

In this work, we investigated the capability of LP models that incorporate numerical literal information and the suitability of the corresponding benchmark datasets. We propose a methodology to create semi-synthetic datasets and a dataset ablation methodology. 

With a semi-synthetic dataset we showed that many models underutilize literal information, even in a setting where the numerical data is crucial for the prediction. We showed that under the established evaluation schema, the performance gains of many models can be attributed to the additional model parameters rather than the models's capabilities to exploit literals.

Future work could investigate real-world KGs regarding their suitability for evaluating numerical LP more deeply. Additionally, developing more challenging synthetic dataset extensions requiring the combination of literal information and graph structure for predicting missing links could offer valuable insights into the potentially more advanced LP models proposed in the future.

\section{Limitations}
We see three limitations of our work: 

i) Our synthetic dataset implements one simple learning goal that requires the model to learn a threshold value to make correct predictions. This learning goal is simple and does not provide any information about the models capabilities in understanding more complex scenarios. More complex learning goals could go beyond numerical literals and could also require to combine information from numerical literals and relational triples. However, we did not investigate complex learning goals, as we believe that if models fail in simple scenarios, they will also fail in more complex ones.

%{\color{red} Relation ablation only applied to one dataset, FB15k-237. Applying it to the others would also be interesting, but requires extensie computational resources for the evalution, as these datasets are significantly larger. But it is possible!}

ii) We did not find a model that consistently shows benefits from the numerical literals provided by the existing benchmark datasets, not even in the relational triples ablation scenario. Therefore, we can not make any conclusions about the value of the literals provided for these datasets. The numerical literals are either not relevant for the prediction task, or the information contained is difficult to use by the existing models.

iii) We exclude the evaluation of Graph Neural Network models, such as R-GCN~\cite{schlichtkrull2018rgcn}, which have the ability to process numerical literals as node features. This decision is based on the absence of published research specifically advocating for these models' application in LP with numerical literal data. 

\section*{Ethical Statement}
We address concerns related to the experimental design and the significance of results of existing methods. Our objective is not to criticize the creators of the models and datasets, but rather to assist the community by providing practical guidance for future research.

In this work, efforts are made to interpret and understand how well-established models respond to changes in literal data. However, it is important to note that explainability methods still encounter challenges in interpreting such models.

All datasets utilized in this research adhere to ethical standards and are obtained from publicly available sources.

\section*{Supplemental Material Statement}
You can find all our source code, datasets, training result logs, and visualization Jupyter Notebooks on GitHub.\footnote{See \url{https://github.com/moritzblum/LiteralEvaluation}.}

\section*{Acknowledgements}
This work was supported by: the Ministry of Culture and Science of the State of North Rhine-Westphalia~(Germany) through SAIL, Grant No. NW21-059A; the Deutsche Forschungsgemeinschaft~(DFG) through the priority program RATIO~(SPP-1999), Grant No. 376059226; the Research Council of Norway through its Centres of Excellence scheme, Integreat -- Norwegian Centre for knowledge-driven machine learning, project number 332645.

\bibliographystyle{splncs04}
\bibliography{custom}

\newpage
\appendix

\section{Dataset Statistics}
\label{sec:datasetstatistics}

Tab.~\ref{tab:datasetstatistics} shows the statistics of the evaluated LP datasets FB15k-237, YAGO3-10k, LitWD48K, and of our Synthetic dataset. 

\begin{table*}[!h]
\setlength{\tabcolsep}{2pt}
\renewcommand{\arraystretch}{1.1}

\caption{Dataset statistics. For LitWD48K, we only consider the attributive triples of type \texttt{xsd:decimal} as described in Section~\ref{datasets}. We show original numbers indicated by  $^*$ in front of the affected numbers. }
\label{tab:datasetstatistics}

\begin{center}
\footnotesize
\begin{tabular}{l|rrrr}
\hline
\textbf{Dataset}                            & \textbf{FB15k-237} & \textbf{YAGO3-10} & \textbf{LitWD48K}   & \textbf{Synthetic} \\ \hline
\# entities ($|\mathcal{E}|$)               & 14,541             & 123,182           & 47,998               & 14,541             \\
\# relations ($|\mathcal{R}_E|$)            & 237                & 37                & 257                  & 237                \\
\# attributes ($|\mathcal{R}_A|$)           & 121                & 5                 & {\scriptsize (*291)} 246             & 1                  \\
\# relational triples  ($|\mathcal{G}_E|$)  & 310,116            & 1,089,040         & 336,745              & 310,116            \\
\# attributive triples ($|\mathcal{G}_A|$)  & 70,257             & 111,406           & {\scriptsize (*324,418)} 148,707     & 14,541             \\ 
\# entities w/o num.                        & 4600               & 31,030            & {\scriptsize (*0)} 8,198             & 0                  \\ \hline
\# train                                    & 272,115            & 1,079,040         & 303,117              & 272,115            \\
\# test                                     & 17,535             & 5,000             & 16,838               & 17,535             \\
\# valid                                    & 20,466             & 5,000             & 16,838               & 20,466             \\ \hline
\end{tabular}
\end{center}
\end{table*}

In FB15k-237, 3/10 of the triples related to the most frequent attributive relations hold IDs for other databases, overall devoting to 6.9\% of the attributive triples in the dataset. Tab.~\ref{tab:attributive_relations_fb} shows the 10 most frequent attributive relations in FB15k-237. 

\begin{table*}[!h]

\centering
\setlength{\tabcolsep}{6pt}
\renewcommand{\arraystretch}{1.1}

\caption{Ten most frequent attributive relations in FB15k-237. The original relation URIs are \texttt{http://rdf.freebase.com/ns/ + relation name}.}
\label{tab:attributive_relations_fb}

\begin{tabular}{lr}
\hline
\textbf{Relation} & \textbf{\# triples} \\ \hline
\textit{topic\_server.population\_number}                       &  52764 \\
\textit{people.person.height\_meters}                           &   2871 \\
\textit{location.location.area}                                 &   2166 \\
\textit{film.film.netflix\_id}                                  &   1883 \\
\textit{organization.organization.date\_founded}                &    844 \\
\textit{user.robert.default\_domain.rated\_film.ew\_rating}     &    739 \\
\textit{location.location.gnis\_feature\_id}                    &    645 \\
\textit{sports.sports\_team.founded}                            &    643 \\
\textit{location.hud\_county\_place.countyplace\_id}            &    568 \\
\textit{tv.tv\_program.episode\_running\_time}                  &    493 \\ \hline
\end{tabular}
\end{table*}

\section{Hyperparameters}
\label{sec:hyperparameters}

We either used the hyperparameters reported as best in the original publications, or performed a hyperparameter optimization in case we re-implemented a model. 

We refrained from performing a new hyperparameter optimization for the models trained on our ablated datasets, as it is common practice to use the same hyperparameters across multiple datasets \cite{kristiadi2019incorporating, ijcai2022p316}. Further, Ruffinelli et al. show that the relative performance difference between various LP model architectures often shrunk through hyperparameter optimization and re-implementation when compared to prior results~\cite{ruffinelli2020you}. 

We use the following hyperparameters:

\paragraph{LiteralE$_{DistMult}$, LiteralE$_{ComplEx}$), KBLN, and MTKGNN} 
We use the best hyperparameters reported in the original publication to ensure a fair comparison: embedding dim. 200, epochs 100, learning rate 0.001, batch size 128, embedding dropout prob. 0.2, and label smoothing 0.1. The same hyperparameters are used across all models and datasets.

\paragraph{TransEA}
We carried out a grid-search hyperparameter optimization: embedding dim. \{50, \textbf{100}\}, learning rate \{0.01, \textbf{0.001}\}, and $\alpha$ \{\textbf{0.1}, 0.2, ..., 0.9\}.\footnote{We did not apply dropout or label smoothing, thereby following \cite{wu2018transea}.}  The best hyperparameters are highlighted. Further, we set: epochs 500, and batch size 128.

\paragraph{KGA}
We used the  Quantile Hierarchy augmentation method, which showed the best results across all models on FB15K-237 according to \cite{ijcai2022p316}. For KGA$_{TuckER}$ we use the hyperparameters: embedding dim. 200, epochs 500, learning rate 0.003, batch size 128, embedding dropout prob. 0.2, and hidden dropout 0.3. For KGA$_{DistMult}$ we use additional label smoothing 0.1. \\

All models are trained for LP as usual. We applied early stopping by monitoring the MRR score on the validation set every three epochs.

\section{Detailed Results}
\label{sec:app_tables}

We report the MR, MRR, and Hits@10 scores (their mean and variance) obtained over three runs for models and datasets either
including original features or including random features in Tab.~\ref{tab:org_vs_rand}.

\begin{figure}[!h]
     \centering
     \begin{subfigure}[b]{\textwidth}
        \centering
         \includegraphics[trim={0 0.4cm 0 0}, width=0.8\textwidth]{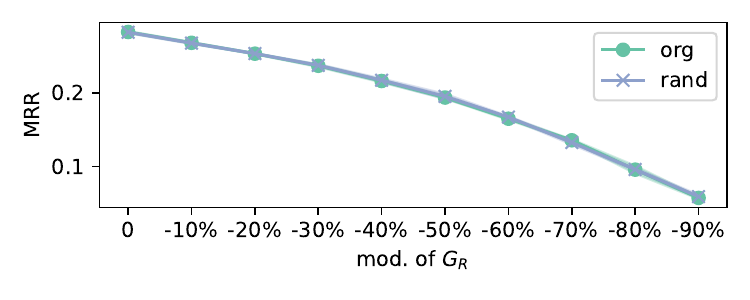}
        \caption{MTKGNN}
        \label{fig:rel_abl_mtkgnn}
     \end{subfigure}

    \hfill
     
     \begin{subfigure}[b]{\textwidth}
        \centering
        \includegraphics[trim={0 0.4cm 0 0}, width=0.8\textwidth]{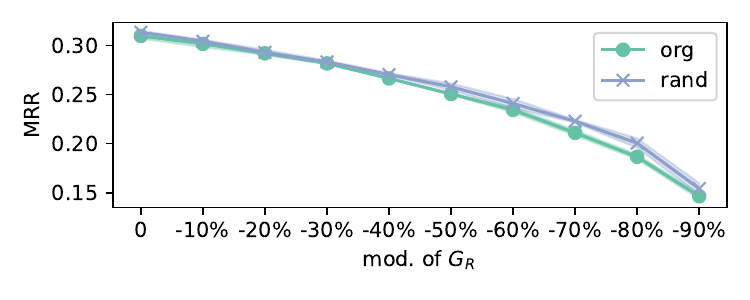}
        \caption{LiteralE$_{DistMult}$}
        \label{fig:rel_abl_literaledistmult}
     \end{subfigure}

    \hfill
     
     \begin{subfigure}[b]{\textwidth}
        \centering
        \includegraphics[trim={0 0.4cm 0 0}, width=0.8\textwidth]{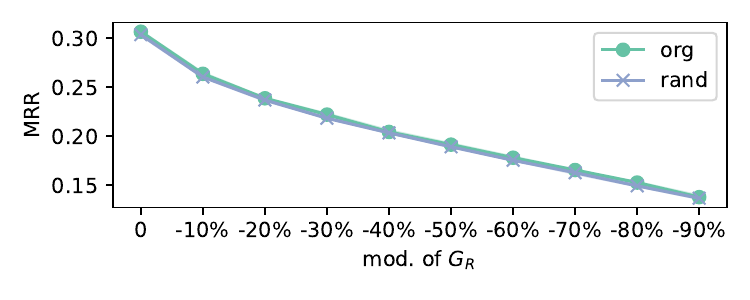}
        \caption{KGA$_{DistMult}$}
        \label{fig:rel_abl_kgadistmult}
     \end{subfigure}
     
     \caption{MRR score after removing some percentage of relational triples from FB15k-237. The models are provided either with the original numerical features or with random features. Mean score and variance are shown across three runs.}
     \label{fig:rel_abl_app}
\end{figure}

Furthermore, the effect of the relational triples ablation from FB15k-237 on MTKGNN and LiteralE$_{DistMult}$ is shown in Fig.~\ref{fig:rel_abl_app}.

\begin{table*}[!h!]
\setlength{\tabcolsep}{6pt}
\renewcommand{\arraystretch}{1.1}

\caption{Comparison of scores of models trained on the datasets provided with the original literal feature versus those trained on datasets provided with random features.}
\label{tab:org_vs_rand}
\begin{center}
\begin{tabular}{l|rll|rll}

\hline
      & \multicolumn{3}{c|}{\textbf{Original features}} & \multicolumn{3}{c}{\textbf{Random features}} \\
\textbf{Model}      & \textbf{MR} & \textbf{MRR} & \textbf{Hits@1} & \textbf{MR}  & \textbf{MRR} & \textbf{Hits@1} \\ \hline

\hline \multicolumn{7}{c}{\textbf{FB15k-237}} \\ \hline 
LiteralE$_{DistMult}$& $285{\scriptstyle \pm 001}$ & $.310{\scriptstyle \pm .003}$ & $.224{\scriptstyle \pm .003}$ & $286{\scriptstyle \pm 008}$ & $.313{\scriptstyle \pm .001}$ & $.229{\scriptstyle \pm .002}$  \\ 
LiteralE$_{ComplEx}$& $429{\scriptstyle \pm 002}$ & $.272{\scriptstyle \pm .000}$ & $.193{\scriptstyle \pm .001}$ & $414{\scriptstyle \pm 022}$ & $.278{\scriptstyle \pm .001}$ & $.198{\scriptstyle \pm .001}$  \\ 
KBLN& $486{\scriptstyle \pm 004}$ & $.295{\scriptstyle \pm .000}$ & $.213{\scriptstyle \pm .001}$ & $618{\scriptstyle \pm 007}$ & $.285{\scriptstyle \pm .002}$ & $.207{\scriptstyle \pm .003}$  \\ 
MTKGNN& $563{\scriptstyle \pm 006}$ & $.282{\scriptstyle \pm .001}$ & $.202{\scriptstyle \pm .001}$ & $575{\scriptstyle \pm 004}$ & $.282{\scriptstyle \pm .001}$ & $.202{\scriptstyle \pm .001}$  \\ 
TransEA& $303{\scriptstyle \pm 002}$ & $.203{\scriptstyle \pm .002}$ & $.132{\scriptstyle \pm .002}$ & $305{\scriptstyle \pm 001}$ & $.200{\scriptstyle \pm .003}$ & $.128{\scriptstyle \pm .002}$  \\ 
KGA$_{TuckER}$& $200{\scriptstyle \pm 004}$ & $.324{\scriptstyle \pm .001}$ & $.234{\scriptstyle \pm .000}$ & $200{\scriptstyle \pm 001}$ & $.322{\scriptstyle \pm .001}$ & $.231{\scriptstyle \pm .001}$  \\ 
KGA$_{DistMult}$& $372{\scriptstyle \pm 007}$ & $.307{\scriptstyle \pm .001}$ & $.221{\scriptstyle \pm .002}$ & $389{\scriptstyle \pm 004}$ & $.304{\scriptstyle \pm .001}$ & $.220{\scriptstyle \pm .002}$  \\ 
\hline \multicolumn{7}{c}{\textbf{YAGO3-10}} \\ \hline 
LiteralE$_{DistMult}$& $1860{\scriptstyle \pm 018}$ & $.479{\scriptstyle \pm .003}$ & $.400{\scriptstyle \pm .004}$ & $1925{\scriptstyle \pm 058}$ & $.470{\scriptstyle \pm .003}$ & $.388{\scriptstyle \pm .004}$  \\ 
LiteralE$_{ComplEx}$& $2086{\scriptstyle \pm 034}$ & $.475{\scriptstyle \pm .002}$ & $.400{\scriptstyle \pm .002}$ & $2303{\scriptstyle \pm 116}$ & $.480{\scriptstyle \pm .004}$ & $.408{\scriptstyle \pm .004}$  \\ 
KBLN& $2850{\scriptstyle \pm 129}$ & $.485{\scriptstyle \pm .023}$ & $.405{\scriptstyle \pm .024}$ & $4243{\scriptstyle \pm 057}$ & $.483{\scriptstyle \pm .001}$ & $.401{\scriptstyle \pm .001}$  \\ 
MTKGNN& $3287{\scriptstyle \pm 048}$ & $.449{\scriptstyle \pm .017}$ & $.362{\scriptstyle \pm .016}$ & $3235{\scriptstyle \pm 094}$ & $.467{\scriptstyle \pm .001}$ & $.379{\scriptstyle \pm .002}$  \\ 
TransEA& $2226{\scriptstyle \pm 041}$ & $.084{\scriptstyle \pm .002}$ & $.048{\scriptstyle \pm .001}$ & $2325{\scriptstyle \pm 055}$ & $.084{\scriptstyle \pm .001}$ & $.049{\scriptstyle \pm .001}$  \\ 
KGA$_{TuckER}$& $1046{\scriptstyle \pm 003}$ & $.497{\scriptstyle \pm .004}$ & $.412{\scriptstyle \pm .005}$ & $1094{\scriptstyle \pm 020}$ & $.492{\scriptstyle \pm .002}$ & $.407{\scriptstyle \pm .002}$  \\ 
KGA$_{DistMult}$& $1554{\scriptstyle \pm 005}$ & $.495{\scriptstyle \pm .003}$ & $.407{\scriptstyle \pm .004}$ & $1696{\scriptstyle \pm 071}$ & $.496{\scriptstyle \pm .002}$ & $.410{\scriptstyle \pm .004}$  \\ 
\hline \multicolumn{7}{c}{\textbf{LitWD48k}} \\ \hline 
LiteralE$_{DistMult}$& $886{\scriptstyle \pm 060}$ & $.326{\scriptstyle \pm .003}$ & $.250{\scriptstyle \pm .003}$ & $875{\scriptstyle \pm 041}$ & $.330{\scriptstyle \pm .004}$ & $.250{\scriptstyle \pm .001}$  \\ 
LiteralE$_{ComplEx}$& $1489{\scriptstyle \pm 109}$ & $.268{\scriptstyle \pm .005}$ & $.200{\scriptstyle \pm .007}$ & $1358{\scriptstyle \pm 067}$ & $.305{\scriptstyle \pm .005}$ & $.238{\scriptstyle \pm .003}$  \\ 
KBLN& $1741{\scriptstyle \pm 042}$ & $.329{\scriptstyle \pm .004}$ & $.246{\scriptstyle \pm .004}$ & $1836{\scriptstyle \pm 044}$ & $.334{\scriptstyle \pm .002}$ & $.262{\scriptstyle \pm .002}$  \\ 
MTKGNN& $3085{\scriptstyle \pm 208}$ & $.289{\scriptstyle \pm .002}$ & $.227{\scriptstyle \pm .003}$ & $2743{\scriptstyle \pm 137}$ & $.297{\scriptstyle \pm .001}$ & $.235{\scriptstyle \pm .001}$  \\ 
TransEA& $947{\scriptstyle \pm 019}$ & $.197{\scriptstyle \pm .003}$ & $.131{\scriptstyle \pm .003}$ & $952{\scriptstyle \pm 032}$ & $.195{\scriptstyle \pm .005}$ & $.129{\scriptstyle \pm .004}$  \\ 
KGA$_{TuckER}$& $353{\scriptstyle \pm 010}$ & $.403{\scriptstyle \pm .001}$ & $.315{\scriptstyle \pm .001}$ & $395{\scriptstyle \pm 007}$ & $.376{\scriptstyle \pm .001}$ & $.293{\scriptstyle \pm .001}$  \\ 
KGA$_{DistMult}$& $504{\scriptstyle \pm 005}$ & $.335{\scriptstyle \pm .001}$ & $.250{\scriptstyle \pm .001}$ & $1422{\scriptstyle \pm 094}$ & $.227{\scriptstyle \pm .004}$ & $.169{\scriptstyle \pm .003}$  \\ 

\hline
\end{tabular}
\end{center}
\end{table*}

\begin{sidewaysfigure*}

  \centering
  \includegraphics[width=\textwidth]{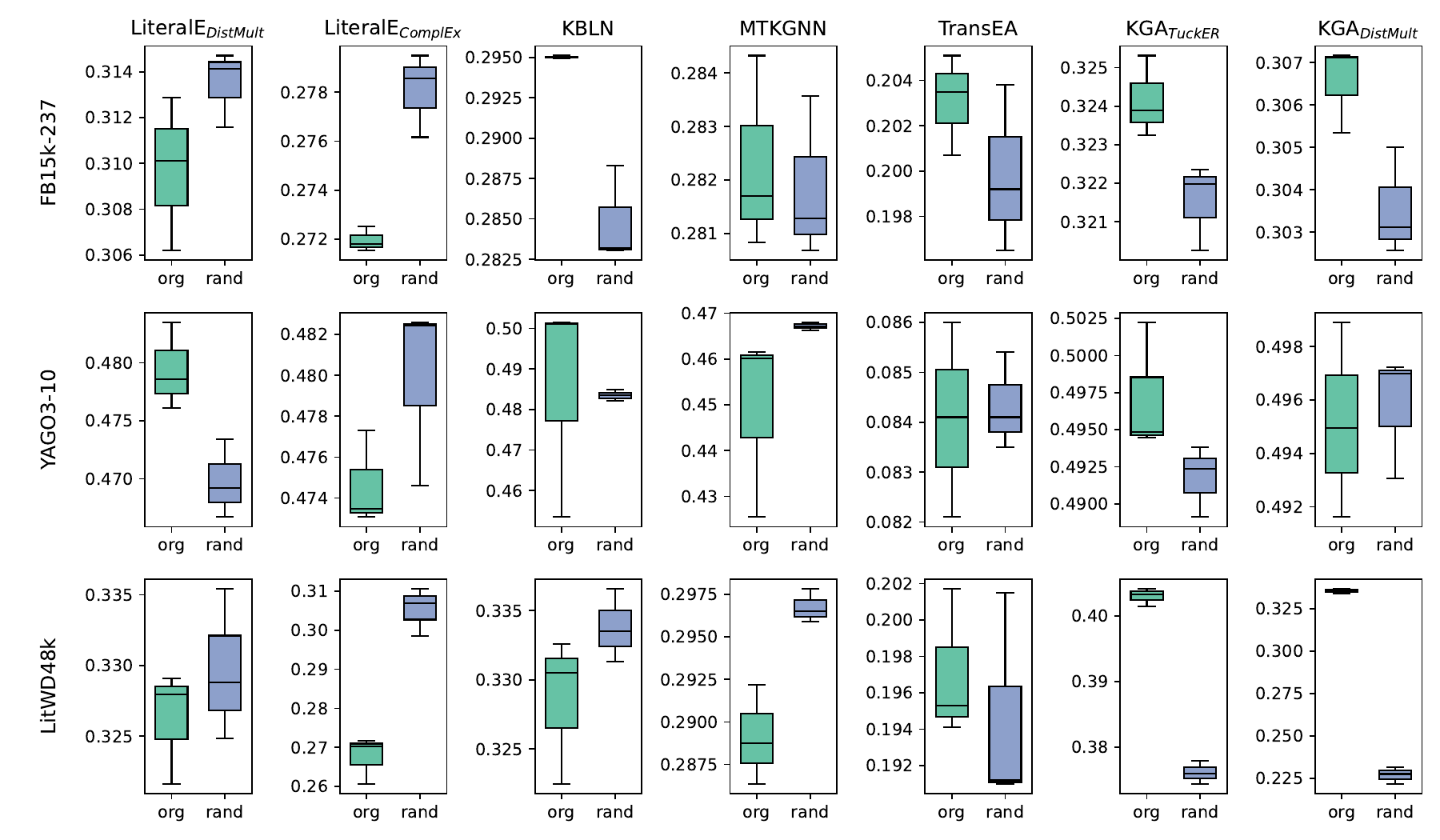}
  \caption{MRR scores obtained over three runs for models and datasets either including original features or including random features. The variance is marginally small and not visually recognizable in the figure.
  }
  \label{fig:orgvsrandl}

\end{sidewaysfigure*}

\section{Further Ablation Experiments: Attributive Value Feature}
\label{sec:app_furtherablation}

In order to provide further insights into the literal features of existing datasets, we apply an additional literal ablation method that allows to investigate whether the concrete literal value is important, or whether only the existence of such an attribute is important, or whether only the existence of an attribute can be taken into account by a model.  

Therefore, we propose an ablation method that removes the concrete literal values but adds literal values that indicate whether the attribute exists. Given $G = G_E \cup G_A$, we create $G' = G_E \cup G_A'$ where $G_A'$ is created as follows: if there exists a value $v$ such that $(e,p,v) \in G_A$, then we add the triple $(e,p,1)$ to $G_A'$.

We evaluate the two models, KGA$_{TuckER}$ and KGA$_{DistMult}$, which exhibit the highest benefits when provided with literals. Tab.~\ref{tab:variations} displays the scores of these models on three datasets. The MRR scores show no consistent trend, but the MR scores are consistently worse for the models that are provided only with the existence of attributes of entities compared to those provided with original or random features. We hypothesize that the lower performance resulting from abstracting the concrete literal values is due to the transformations that transform attributive triples into relational triples, which create a disadvantageous graph structure where most entities are connected to the two entities representing the literal values "0" and "1".

\begin{table*}[]
\setlength{\tabcolsep}{6pt}
\renewcommand{\arraystretch}{1.1}

\caption{Comparison of models trained and evaluated on datasets, each subjected to all proposed literal ablations. The model that uses the literals provided with the dataset is named \textit{original}. The models that use only the relation type and not the concrete literal value are are named \textit{relation type}. }
\label{tab:variations}
\begin{center}

\begin{tabular}{l|l||lllll}
\hline
& \multicolumn{1}{l||}{\textbf{features}} & \textbf{MR} & \textbf{MRR} & \textbf{Hits@1} & \textbf{Hits@3} & \textbf{Hits@10} \\
\hline \multirow{8}{*}{\rotatebox[origin=c]{90}{FB15k-237}} 
& \multicolumn{6}{c}{\textbf{KGA$_{TuckER}$}} \\ 
 \cline{2-7} 
& original & $200{\scriptstyle \pm 004}$ & $.324{\scriptstyle \pm .001}$ & $.234{\scriptstyle \pm .000}$ & $.355{\scriptstyle \pm .002}$ & $.508{\scriptstyle \pm .001}$ \\ 
& random & $200{\scriptstyle \pm 001}$ & $.322{\scriptstyle \pm .001}$ & $.231{\scriptstyle \pm .001}$ & $.354{\scriptstyle \pm .000}$ & $.505{\scriptstyle \pm .001}$ \\ 
& relation type & $217{\scriptstyle \pm 002}$ & $.319{\scriptstyle \pm .001}$ & $.229{\scriptstyle \pm .001}$ & $.351{\scriptstyle \pm .001}$ & $.502{\scriptstyle \pm .001}$ \\ 
\cline{2-7} 
& \multicolumn{6}{c}{\textbf{KGA$_{DistMult}$}} \\ 
 \cline{2-7} 
& original & $372{\scriptstyle \pm 007}$ & $.307{\scriptstyle \pm .001}$ & $.221{\scriptstyle \pm .002}$ & $.336{\scriptstyle \pm .001}$ & $.478{\scriptstyle \pm .001}$ \\ 
& random & $389{\scriptstyle \pm 004}$ & $.304{\scriptstyle \pm .001}$ & $.220{\scriptstyle \pm .002}$ & $.333{\scriptstyle \pm .001}$ & $.472{\scriptstyle \pm .001}$ \\ 
& relation type & $401{\scriptstyle \pm 011}$ & $.303{\scriptstyle \pm .002}$ & $.218{\scriptstyle \pm .002}$ & $.333{\scriptstyle \pm .001}$ & $.474{\scriptstyle \pm .005}$ \\ 
\hline \multirow{8}{*}{\rotatebox[origin=c]{90}{YAGO3-10}} 
& \multicolumn{6}{c}{\textbf{KGA$_{TuckER}$}} \\ 
 \cline{2-7} 
& original & $1046{\scriptstyle \pm 003}$ & $.497{\scriptstyle \pm .004}$ & $.412{\scriptstyle \pm .005}$ & $.546{\scriptstyle \pm .005}$ & $.651{\scriptstyle \pm .001}$ \\ 
& random & $1094{\scriptstyle \pm 020}$ & $.492{\scriptstyle \pm .002}$ & $.407{\scriptstyle \pm .002}$ & $.539{\scriptstyle \pm .002}$ & $.646{\scriptstyle \pm .002}$ \\ 
& relation type & $1320{\scriptstyle \pm 064}$ & $.521{\scriptstyle \pm .005}$ & $.439{\scriptstyle \pm .005}$ & $.570{\scriptstyle \pm .003}$ & $.671{\scriptstyle \pm .005}$ \\ 
\cline{2-7} 
& \multicolumn{6}{c}{\textbf{KGA$_{DistMult}$}} \\ 
 \cline{2-7} 
& original & $1554{\scriptstyle \pm 005}$ & $.495{\scriptstyle \pm .003}$ & $.407{\scriptstyle \pm .004}$ & $.543{\scriptstyle \pm .002}$ & $.661{\scriptstyle \pm .001}$ \\ 
& random & $1696{\scriptstyle \pm 071}$ & $.496{\scriptstyle \pm .002}$ & $.410{\scriptstyle \pm .004}$ & $.543{\scriptstyle \pm .002}$ & $.655{\scriptstyle \pm .002}$ \\ 
& relation type & $1755{\scriptstyle \pm 061}$ & $.509{\scriptstyle \pm .002}$ & $.423{\scriptstyle \pm .002}$ & $.559{\scriptstyle \pm .004}$ & $.668{\scriptstyle \pm .002}$ \\ 
\hline \multirow{8}{*}{\rotatebox[origin=c]{90}{LitWD48k}} 
& \multicolumn{6}{c}{\textbf{KGA$_{TuckER}$}} \\ 
 \cline{2-7} 
& original & $353{\scriptstyle \pm 010}$ & $.403{\scriptstyle \pm .001}$ & $.315{\scriptstyle \pm .001}$ & $.435{\scriptstyle \pm .002}$ & $.584{\scriptstyle \pm .000}$ \\ 
& random & $395{\scriptstyle \pm 007}$ & $.376{\scriptstyle \pm .001}$ & $.293{\scriptstyle \pm .001}$ & $.403{\scriptstyle \pm .002}$ & $.545{\scriptstyle \pm .002}$ \\ 
& relation type & $672{\scriptstyle \pm 013}$ & $.386{\scriptstyle \pm .001}$ & $.303{\scriptstyle \pm .001}$ & $.418{\scriptstyle \pm .000}$ & $.556{\scriptstyle \pm .001}$ \\ 
\cline{2-7} 
& \multicolumn{6}{c}{\textbf{KGA$_{DistMult}$}} \\ 
 \cline{2-7} 
& original & $504{\scriptstyle \pm 005}$ & $.335{\scriptstyle \pm .001}$ & $.250{\scriptstyle \pm .001}$ & $.360{\scriptstyle \pm .002}$ & $.513{\scriptstyle \pm .003}$ \\ 
& random & $1422{\scriptstyle \pm 094}$ & $.227{\scriptstyle \pm .004}$ & $.169{\scriptstyle \pm .003}$ & $.243{\scriptstyle \pm .005}$ & $.339{\scriptstyle \pm .006}$ \\ 
& relation type & $786{\scriptstyle \pm 015}$ & $.359{\scriptstyle \pm .001}$ & $.268{\scriptstyle \pm .001}$ & $.388{\scriptstyle \pm .002}$ & $.557{\scriptstyle \pm .003}$ \\ 

\hline
\end{tabular}
\end{center}
\end{table*}

\end{document}